\documentclass[a4paper]{cas-dc}

\usepackage[authoryear]{natbib}
\usepackage{amsmath,amssymb,amsfonts}
\usepackage{graphicx}
\usepackage{textcomp}
\usepackage{booktabs}
\usepackage{multirow}
\usepackage{url}
\usepackage{caption}
\usepackage{makecell}

\begin{document}
\let\WriteBookmarks\relax
\shorttitle{Memory-Space Adaptation for Medical Image Segmentation}
\shortauthors{Chen et al.}

\title[mode=title]{From Weight-Space to Memory-Space Adaptation:
A Unified Framework for Medical Image Segmentation}

\author[1]{Bowen Chen}
\author[2]{Qiaohui Gao}
\author[1]{Shaowen Wan}
\author[3]{Ruiyu Yan}
\author[4]{Shanhui Sun}
\author[5]{Wei Liu}
\author[6]{Xiang Li}
\author[7]{Tianming Liu}
\author[1]{Lin Zhao}
\cormark[1]
\ead{lin.zhao.1@njit.edu}
\cortext[1]{Corresponding author}

\affiliation[1]{organization={Department of Biomedical Engineering, New Jersey Institute of Technology},city={Newark},state={NJ},postcode={07102},country={USA}}
\affiliation[2]{organization={College of Engineering, Northeastern University},city={Boston},state={MA},postcode={02115},country={USA}}
\affiliation[3]{organization={Tandon School of Engineering, New York University},city={Brooklyn},state={NY},postcode={11201},country={USA}}
\affiliation[4]{organization={United Imaging Intelligence},city={Burlington},state={MA},postcode={01803},country={USA}}
\affiliation[5]{organization={Department of Radiation Oncology, Mayo Clinic},city={Scottsdale},state={AZ},postcode={85259},country={USA}}
\affiliation[6]{organization={Department of Radiology, Massachusetts General Hospital and Harvard Medical School},city={Boston},state={MA},postcode={02115},country={USA}}
\affiliation[7]{organization={School of Computing, University of Georgia},city={Athens},state={GA},postcode={30602},country={USA}}

\begin{abstract}
We introduce \emph{memory-space adaptation}, a paradigm for medical image segmentation in which adaptation is realized through lightweight memory operations rather than repeated model-parameter optimization. Existing approaches primarily operate in weight space, requiring parameter or task-specific module updates for new tasks or distribution shifts and incurring substantial communication costs in federated settings. In memory-space adaptation, supervised task adaptation, few-shot learning, and feedback-driven online adaptation are realized through the construction and evolution of memory within a shared representation space, unifying learning and adaptation settings that are conventionally treated separately while the pretrained model remains fixed. We instantiate this paradigm using the native memory space of a frozen SAM2-family backbone and develop \emph{MemSeg-Agent}, which realizes these adaptation regimes through static, few-shot, and working memories coordinated by an agentic read--write controller for memory retrieval, composition, update, and consolidation. Across four public datasets spanning multiple imaging modalities and anatomical regions, MemSeg-Agent achieves competitive supervised segmentation across annotation budgets and substantially improves cross-domain performance through online adaptation. In federated settings, memory-only exchange reduces per-round communication by approximately 74.3$\times$ while maintaining comparable segmentation performance. These results establish memory-space adaptation as a promising paradigm for adaptive medical image segmentation and suggest a broader direction toward AI systems that continuously acquire, consolidate, and reuse experience without repeatedly updating model parameters.
\end{abstract}



\maketitle

\section{Introduction}
\label{sec:introduction}

 Medical image segmentation is crucial for image-based diagnosis, treatment planning, and image-guided intervention. Over the past decade, deep learning based method have achieved strong segmentation performance across a wide range of anatomical structures and imaging modalities \cite{litjens2017survey,azad2024medical}. More recently, vision foundation models, such as DINO and SAM, have provided highly transferable representations and have been increasingly adapted for medical image segmentation, offering improved generalization and data efficiency\cite{oquab2023dinov2,ravi2024sam,ma2024medsam,zhao2025retrieval}. Despite these advancements, clinical deployment often requires further adaptations to distribution shifts arising from differences in institutions, scanners, acquisition protocols, downstream tasks, and patient populations\cite{ovadia2019can,kelly2019key}. Current adaptation strategies primarily encode task- and domain-specific information through updates to model weights, including full fine-tuning, parameter efficient fine-tuning (PEFT), and sequential parameter updates \cite{zhang2025peft,wang2025parameterefficientfinetuninglargemodels}. This weight-space organization ties adaptation to repeated parameter optimization, increasing model-maintenance complexity as deployment environments evolve and introducing substantial communication overhead under Federated Learning (FL) where model updates are exchanged across institutions \cite{zhuang2025foundationmodelmeetsfederated,liu2024fedfms,zhang2025fedsca}. As segmentation models grow in scale and deployment environments become increasingly dynamic and heterogeneous, how adaptation knowledge is represented, stored, and efficiently updated stands out as a central challenge in building scalable and practically deployable image segmentation systems.

Viewed more fundamentally, adaptation is a problem of learning: how can a model efficiently acquire knowledge from limited experience, retain and internalize that knowledge over time, and reuse it to respond to changing environments? Few-shot learning, continual learning, and online adaptation address different facets of this broader problem. Few-shot learning focuses on acquiring useful knowledge from limited examples, continual learning on integrating new experience over time without losing previously acquired knowledge, and online adaptation on incorporating feedback or newly available information during deployment to respond to changing data distributions and task conditions \cite{zhao2025retrieval,gonzalez2023lifelongnnunet,ranem2024unclesam,chen2024each}. The deeper limitation is that learning remains compartmentalized across these paradigms, such that experience acquired in one setting does not naturally accumulate into a persistent knowledge state that can support learning in another. A more general solution calls for a common learning substrate through which experience acquired under different conditions can be progressively integrated, retained, and reused.

Biological learning provides a compelling perspective on this problem. The brain distributes learning across complementary memory systems operating at different timescales. The hippocampal system rapidly encodes specific experiences, while neocortical networks integrate information more gradually into distributed representations that capture regularities and support generalization \cite{mcclelland1995complementary,kumaran2016learning}. Through reactivation and replay, recent experiences can be progressively integrated with existing knowledge, supporting the coexistence of rapid learning and long-term stability \cite{wilson1994reactivation,carr2011hippocampal,diekelmann2010memory}. Retrieval further reactivates established memories and can return them to a plastic state, allowing prior knowledge to be updated in light of new experience \cite{nader2009single}. From this perspective, memory functions as an active learning substrate that connects knowledge acquisition, retention, and reuse. Recent advances in agentic memory further demonstrate that artificial systems can explicitly store, retrieve, update, and consolidate experience through memory during interaction \cite{yu2026agenticmemorylearningunified,behrouz2024titanslearningmemorizetest,xu2025amem}. Inspired by these principles, we introduce \emph{memory-space adaptation}, a learning paradigm in which post-training adaptation is realized through the evolution of memory rather than repeated modification of model parameters. By making memory the common substrate of adaptation, experience acquired from different learning conditions can accumulate within a shared representational space and be reused for subsequent tasks.

To instantiate this memory-space formulation, we develop MemSeg-Agent using the native memory interface of SAM2, which provides a pretrained pathway for encoding prior observations and conditioning future predictions through memory \cite{ravi2024sam,zhao2025retrieval}. MemSeg-Agent represents learning through three complementary memory states: static memory captures stable task knowledge, few-shot memory incorporates newly available labeled experience, and working memory consolidates online feedback into evolving task experience. A lightweight read--write controller coordinates memory retrieval, composition, update and consolidation while the SAM2-family backbone remains fixed. To validate MemSeg-Agent across diverse real-world conditions, we conduct experiments on four public datasets spanning multiple imaging modalities and anatomical regions, each designed to probe a distinct deployment scenario. When supervised labels are available, static memory encodes task-specific priors without any backbone modification, achieving competitive segmentation across varying annotation budgets. As models encounter distribution shift during clinical deployment, working memory absorbs knowledge from clinician corrections and substantially outperforms established continual medical segmentation approaches, including UnCLeSAM and Lifelong nnU-Net~\cite{ranem2024unclesam,gonzalez2023lifelongnnunet}. The same lightweight memory representation also supports few-shot adaptation when labeled data is scarce, and reduces federated per-round communication by approximately $\sim$74.3$\times$ ($\sim$98.65\%) by exchanging only compact memory variables rather than full model parameters. Together, we demonstrate that a unified memory space can serve as a common substrate for knowledge acquisition, adaptation, and reuse throughout the model lifecycle.

Our contributions are summarized as follows:
\begin{itemize}

    \item We introduce \emph{memory-space adaptation} as a unified learning paradigm for medical image segmentation, shifting adaptation from repeated parameter updates to the evolution of explicit memories. This formulation provides a common learning substrate for settings conventionally treated separately, including supervised learning, few-shot learning, continual learning, and federated learning.

    \item We develop \emph{MemSeg-Agent}, which instantiates this paradigm on a frozen SAM2-family backbone through static, few-shot, and working memories that represent stable task knowledge, limited new experience, and continuously acquired feedback within a shared representation space, allowing knowledge acquired under different learning conditions to naturally accumulate and be reused across inference steps.

    \item We introduce an agentic read--write memory mechanism that governs how experience is retrieved, composed, updated, and consolidated over time. By transforming newly acquired supervision and clinician feedback into persistent memory, the framework enables knowledge to evolve while preserving a stable foundation-model prior.

    \item We systematically validate memory-space adaptation across multiple learning settings and four public datasets, demonstrating competitive segmentation, superior cross-domain adaptation, and approximately $74.3\times$ lower federated communication. These results establish memory as a viable substrate for unified post-training learning and suggest a broader direction toward adaptive AI systems that evolve through memory rather than repeated weight modification.
    
\end{itemize}

\section{Related Work}
\label{sec:related_work}

\subsection{Foundation Models for Medical Image Segmentation}


Vision foundation models provide two complementary components relevant to medical segmentation. DINOv2 and DINOv3 learn transferable visual representations through large-scale self-supervised pretraining, producing spatially dense features that generalize across imaging domains without task-specific supervision~\cite{oquab2023dinov2,simeoni2025dinov3}. Beyond pretraining objectives, these dense representations have been directly adopted as frozen visual encoders for downstream segmentation: Li \textit{et al.} integrate a frozen DINOv3 backbone into an encoder–decoder architecture that achieves competitive performance across seven medical imaging modalities~\cite{li2025dino}. Zhang \textit{et al.} adapt vision foundation models as efficient encoders for real-time ultrasound image segmentation~\cite{zhang2025adapting}. And Ayzenberg \textit{et al.} augment a prototypical network with a frozen DINOv2 encoder to generate coarse segmentation priors for SAM-guided one-shot adaptation~\cite{ayzenberg2025protosam}. SAM and SAM2 instead provide promptable segmentation architectures, with SAM2 introducing a native memory mechanism for conditioning predictions on previously encoded observations~\cite{kirillov2023segment,ravi2024sam}. Medical variants such as MedSAM and MedSAM2 adapt these models through domain-specific training, while parameter-efficient approaches reduce the number of updated model parameters~\cite{ma2024segment,ma2025medsam2,wang2025parameterefficientfinetuninglargemodels,zhang2025peft}. Zhao \textit{et al.} further combined both components, encoding support images through a frozen DINOv2 retrieval encoder and storing the resulting representations in the SAM2 memory pathway for training-free few-shot segmentation~\cite{zhao2025retrieval}. 
However, existing approaches primarily achieve task adaptation through parameter optimization, prompt-based conditioning, or retrieval of annotated support examples. In contrast to retrieval-based methods that use memory primarily for few-shot inference, MemSeg-Agent treats the native SAM2 memory space as a persistent adaptation state. Specifically, it integrates learnable static memory for supervised adaptation, retrieved few-shot memory, and feedback-driven working memory within a unified representation space, while keeping the pretrained segmentation backbone fixed.

\subsection{Continual adaptation.}

Continual medical segmentation aims to acquire new task- or domain-specific knowledge while limiting catastrophic forgetting. Gonzalez \textit{et al.} introduced Lifelong nnU-Net as a standardized framework that integrates representative regularization-, distillation-, and rehearsal-based continual learning strategies into nnU-Net~\cite{gonzalez2023lifelongnnunet}. Ranem \textit{et al.} developed UnCLe SAM to investigate continual adaptation of SAM-based medical segmentation under sequential tasks~\cite{ranem2024unclesam}. Chen \textit{et al.} proposed VPTTA for continual test-time adaptation, freezing the segmentation model while learning an image-specific visual prompt and reusing historical prompts through a memory bank~\cite{chen2024each}. Ranem \textit{et al.} further introduced NCAdapt, which employs lightweight domain-specific neural cellular automata components to accommodate sequential domain shifts~\cite{ranem2025ncadapt}, while Wang \textit{et al.} proposed CA-SAM with task-specific alignment layers and automatic routing to preserve previously acquired capabilities~\cite{wang2026continual}. These methods incorporate new domain knowledge primarily through parameter updates, optimized prompts, or task-specific modules. Although VPTTA also maintains a frozen segmentation model and a memory bank, its adaptation relies on learned visual prompts. In contrast, MemSeg-Agent incorporates clinician corrections directly into working memory through a read--write controller. Meanwhile, supervised static memory remains fixed as a stable source-domain prior, allowing new experience to accumulate alongside previously learned knowledge.

\subsection{Federated adaptation.}
Federated learning enables multiple institutions to collaboratively train segmentation models without exchanging raw patient data~\cite{sheller2020federated}. However, the increasing size of foundation models makes repeated transmission of model updates a substantial systems bottleneck. Liu \textit{et al.} introduced FedFMS to federate SAM-based medical segmentation and used Medical SAM Adapter in FedMSA to reduce training and communication overhead~\cite{liu2024fedfms}. Zhang \textit{et al.} further proposed FedSCA, combining parameter-efficient local adaptation with partial adapter transmission to reduce communication under heterogeneous medical data~\cite{zhang2025fedsca}. Although such PEFT-based strategies substantially reduce the number of transmitted parameters, the shared state remains tied to model or adapter updates and must be repeatedly communicated during federated optimization. Our federated setting instead exchanges only compact task-specific static-memory variables while keeping the foundation-model backbone fixed at every client, treating federated learning as a communication-efficient application of memory-space adaptation rather than a new federated optimization algorithm.

\subsection{Agentic Memory}
Memory-augmented learning stores past examples, prototypes, or feature representations for subsequent retrieval, while recent agentic memory methods further emphasize persistent memory management during interaction. Behrouz \textit{et al.} introduced Titans to learn and update long-term memory at test time~\cite{behrouz2024titanslearningmemorizetest}, Xu \textit{et al.} proposed A-MEM to dynamically organize and link accumulated experiences~\cite{xu2025amem}, and Yu \textit{et al.} developed AgeMem to jointly manage long- and short-term memory through learnable memory operations~\cite{yu2026agenticmemorylearningunified}. In medical applications, Fallahpour \textit{et al.} developed MedRAX to coordinate specialized tools for chest X-ray reasoning~\cite{fallahpour2025medrax}, while Jia \textit{et al.} incorporated agentic memory with retrieval and evidence grounding for medical question answering~\cite{jia2026agentic}. These studies primarily operate over linguistic context, retrieved evidence, or tool-use trajectories, leaving the role of persistent agentic memory for dense spatial prediction largely unexplored. MemSeg-Agent transfers this read--write memory paradigm to segmentation by treating SAM2-compatible spatial representations as the persistent and editable memory state, through which static, few-shot, and working memories are selectively retrieved, composed, and consolidated to progressively accumulate segmentation experience across deployment.


\section{Method}
\label{sec:method}

\label{sec:overview}

\begin{figure*}[!t]
\centering
\includegraphics[width=\textwidth]{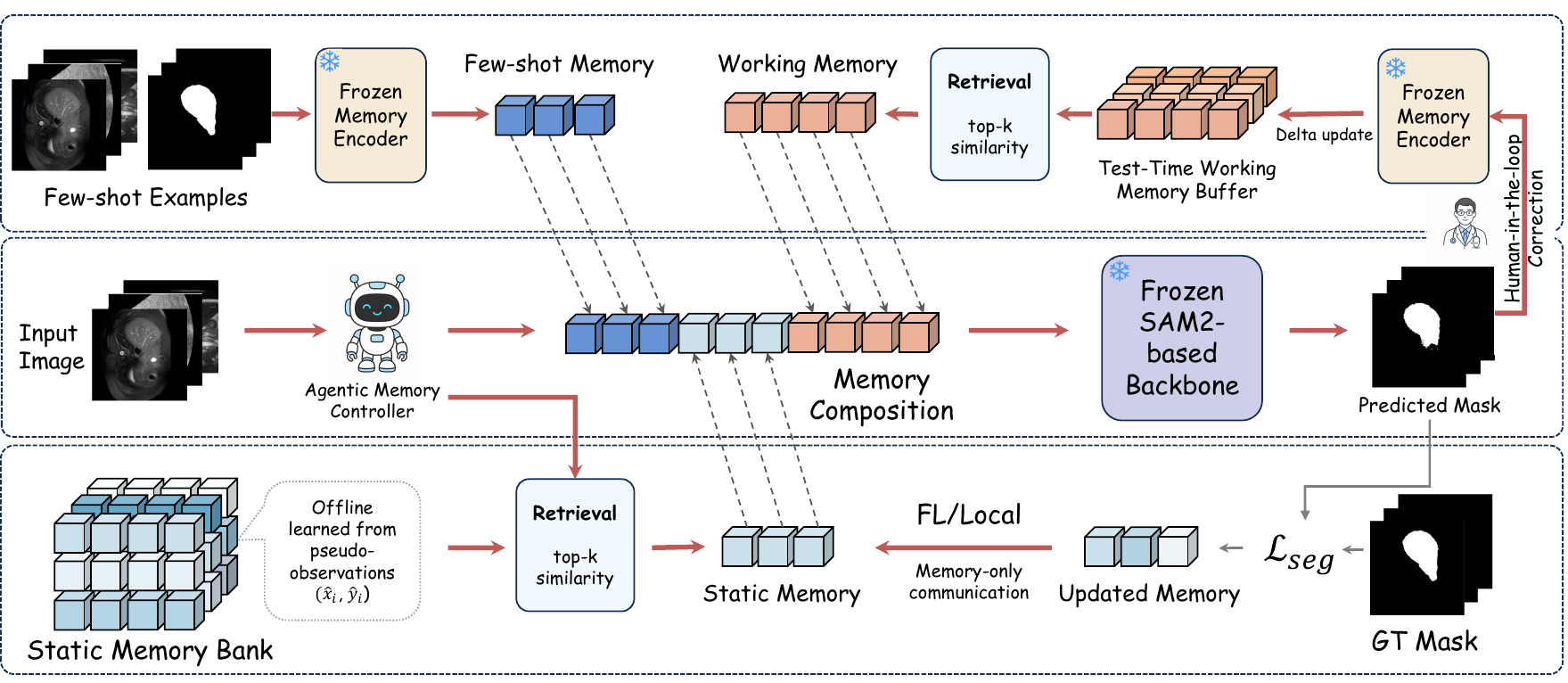}
\caption{Overview of the MemSeg-Agent framework. 
\textbf{Top:} Few-shot and test-time working-memory pathways. Few-shot examples are encoded into reusable memory representations, while human corrections during deployment are encoded as delta updates to a test-time working-memory buffer and retrieved according to similarity. 
\textbf{Middle:} Reference inference pathway. Given an input image, the agentic memory controller retrieves and composes relevant memories from different sources to condition the frozen SAM2-based segmentation backbone for mask prediction. 
\textbf{Bottom:} Static-memory construction and federated/local training. Static memories are learned offline from pseudo-observations and retrieved for inference, while their representations can be optimized  with memory-only communication under the segmentation objective.}
\label{fig:fig1}
\end{figure*}

\subsection{Problem Formulation}

Let $\mathcal{D}=\{(x_i,y_i)\}_{i=1}^{N}$ denote a medical image segmentation dataset, where $x_i$ is an input image and $y_i\in\{0,1\}^{H\times W}$ is the corresponding binary segmentation mask. Each anatomical structure is treated as an independent binary segmentation task with a class-specific memory bank. We consider a pretrained SAM2-family segmentation model $f_{\theta}$, where $\theta$ denotes the parameters of the image encoder, memory encoder, memory attention, prompt encoder, and mask decoder. Conventional adaptation modifies all or a subset of $\theta$ to encode task- or domain-specific knowledge. In contrast, MemSeg-Agent fixes the pretrained model parameters at $\theta_0$ and represents adaptation through an external memory state,
\begin{equation}
    \hat{y}_t =
    f_{\theta_0}
    \left(
        x_t,\varnothing;
        \mathcal{M}^{\mathrm{read}}_t
    \right),
    \qquad
    \theta_t=\theta_0 ,
    \label{eq:memory_space_formulation}
\end{equation}
where $\varnothing$ denotes the empty prompt and $\mathcal{M}^{\mathrm{read}}_t$ denotes the memory selected to condition the prediction of case $t$. We represent the available adaptation state as

\begin{equation}
    \mathbf{M}_t =
    \left(
        \mathcal{M}^{s},
        \mathcal{M}^{f},
        \mathcal{M}^{w}_t
    \right),
    \label{eq:memory_state}
\end{equation}

where $\mathcal{M}^{s}$ is supervised static memory, $\mathcal{M}^{f}$ is retrieval-based few-shot memory, and $\mathcal{M}^{w}_t$ is the test-time working-memory buffer. Depending on the adaptation setting, one or more memory sources may be available. An Agentic Memory Controller selects and composes query-relevant entries from these sources to construct $\mathcal{M}^{\mathrm{read}}_t$.

During sequential deployment, only the working-memory state evolves. For case $t$, prediction is first performed using memories available before the case is processed. If the clinician subsequently provides a corrected mask $y_t^{*}$, the corrected image--mask pair is incorporated into working memory through a memory-update operator $\mathcal{U}$:
\begin{equation}
    \mathcal{M}^{w}_t =
    \begin{cases}
        \mathcal{U}
        \left(
            \mathcal{M}^{w}_{t-1},
            x_t,
            y_t^{*}
        \right),
        & c_t = 1, \\[3pt]
        \mathcal{M}^{w}_{t-1},
        & c_t = 0,
    \end{cases}
    \label{eq:working_memory_update}
\end{equation}
where $c_t\in\{0,1\}$ indicates whether a correction is provided. Importantly, $y_t^{*}$ is introduced only after $\hat{y}_t$ has been evaluated; therefore, the updated memory can affect only subsequent cases $t'>t$. Memory-space adaptation is consequently characterized by an evolving memory state while the segmentation backbone remains unchanged over time.

\subsection{Framework Overview}

We propose \emph{MemSeg-Agent}, a memory-augmented segmentation framework built upon a frozen SAM2-family backbone (Fig.~\ref{fig:fig1}), where adaptation is performed entirely in memory space rather than through repeated model-parameter updates. The framework consists of three complementary memory sources: \emph{static memory} learned from supervised data to encode stable task-specific priors, \emph{few-shot memory} constructed directly from labeled support examples, and test-time \emph{working memory} that accumulates deployment-time corrections. These memory sources share a common SAM2-compatible representation and are coordinated by an \emph{Agentic Memory Controller}, which retrieves, composes, and updates memory according to the current input and deployment state. The following sections describe static-memory construction, few-shot memory, test-time working-memory adaptation, and the read--write controller in detail.

\subsection{Static Memory via Supervision}
\label{sec:static_memory}

SAM2 defines a memory representation space through its pretrained memory encoder. Given an image--mask pair $(x_i,y_i)$, the frozen image encoder $\mathcal{E}_{I}$ and memory encoder $\mathcal{E}_{m}$ produce a SAM2-compatible memory:
\begin{equation}
    m_i=\mathcal{E}_{m}\bigl(\mathcal{E}_{I}(x_i),y_i\bigr).
    \label{eq:memory_encoding}
\end{equation}

To learn task-specific priors without modifying the backbone, we parameterize static memory using learnable pseudo-observations
$z_i=(\hat{x}_i,\hat{y}_i)$. A pseudo-observation is a learnable image--mask prototype rather than a real sample or pseudo-label. It is passed through the frozen observation-to-memory pathway to preserve compatibility with SAM2 memory attention. Given a supervised dataset $\mathcal{D}$, only the pseudo-observations are optimized:
\begin{equation}
    \min_{\{z_i\}_{i=1}^{N}}
    \mathcal{L}_{\mathrm{seg}}
    \left(
        f_{\theta}
        \left(
            x,\varnothing;
            \{\mathcal{E}_{m}(\mathcal{E}_{I}(\hat{x}_i),\hat{y}_i)\}_{i=1}^{N}
        \right),
        y
    \right),
    \qquad \theta\ \mathrm{fixed},
    \label{eq:static_memory}
\end{equation}
where $N$ is fixed to the memory-bank size. We use slice-wise binary segmentation with Dice Loss plus an auxiliary
objectness BCE on foreground-presence logits
\begin{equation}
    \mathcal{L}_{\mathrm{seg}}
    =
    \mathcal{L}_{\mathrm{Dice}}(\hat{y},y)
    +
    0.3\mathcal{L}_{\mathrm{BCE}}(\hat{o},o),
    \label{eq:seg_loss}
\end{equation}
where $o$ and $\hat{o}$ denote the foreground-presence label and corresponding objectness prediction.

In federated settings, each client optimizes the same static-memory variables on local data, and the server consolidates them via FedAvg aggregation. Because only these compact memory variables are transmitted, the frozen foundation model stays local and unchanged.

\subsection{Few-shot Memory}
\label{sec:fewshot_memory}

When labeled data is scarce, directly learning static memory may overfit. Retrieval-augmented few-shot segmentation~\cite{zhao2025retrieval} can be integrated into the same memory interface. Given a support set
$\mathcal{S}=\{(x_i^{s},y_i^{s})\}_{i=1}^{K}$, each support pair is encoded into memory through the frozen image and memory encoders.

A frozen retrieval encoder ranks support examples according to their similarity to the query image. The top-$k$ corresponding memories are then used to condition the frozen backbone. Few-shot memory is therefore non-parametric and requires neither fine-tuning nor additional optimization.

\subsection{Test-time Working Memory}
\label{sec:working_memory}

Static memory provides a stable supervised prior, which however may become insufficient when the deployment distribution differs substantially from the training domain. We therefore introduce test-time working memory to convert clinician feedback into reusable target-domain experience. Unlike conventional test-time adaptation, the segmentation backbone is not optimized during deployment. This adaptation process is activated only when a clinician inspects the current prediction and provides an explicit correction, which is then encoded and incorporated into working memory for future inference.

For each target case $x_t$, MemSeg-Agent first produces a prediction using only the memory state available before case $t$ is processed. The clinician then determines whether the segmentation is satisfactory. If a correction is required, a corrected annotation $y_t^{*}$ is provided and passed to the memory controller for subsequent memory storage and consolidation. Importantly, the correction is introduced only after the prediction for case $t$ has been evaluated and therefore cannot improve the result of the same case. The updated working memory can consequently influence only future cases $t'>t$, yielding a strict prequential evaluation protocol that avoids information leakage from current-case corrections.

We denote the clinician correction decision by $c_t\in\{0,1\}$. In prospective deployment, $c_t$ is determined entirely by the clinician and does not require access to ground truth. For retrospective evaluation, we simulate this interaction using a predefined oracle criterion:
\begin{equation}
c_t =
\mathbb{I}
\left[
\mathrm{Dice}(\hat{y}_t,y_t)<\tau_{\mathrm{corr}}
\right],
\end{equation}
where $\tau_{\mathrm{corr}}$ is a predefined performance threshold used to
simulate clinician dissatisfaction in retrospective evaluation. When $c_t=1$, the ground-truth mask is used as a surrogate clinician-corrected annotation in the retrospective experiments. The threshold therefore serves only to simulate clinician dissatisfaction during retrospective evaluation; it is not an automatic gate required during real deployment.

After a correction is provided, the resulting observation is incorporated into the working-memory buffer according to the read--write controller described in Section~\ref{sec:controller}. This buffer persists across sequential target cases and can be retained after deployment as a domain-specific memory bank for future inference. In this way, each clinician intervention is transformed into reusable target-domain experience that can accumulate over time and improve subsequent cases, allowing adaptation knowledge to evolve entirely within the external memory state without gradient-based updates to the segmentation backbone.

\subsection{Agentic Memory Controller}
\label{sec:controller}

The Agentic Memory Controller governs how available memory sources are retrieved and composed for prediction, as well as how newly corrected observations are incorporated into persistent working memory. It does not determine whether a clinician correction is required for a given case. Instead, the controller implements the read--write policy that coordinates static, few-shot, and test-time working memories and manages their interaction across a sequential stream of target cases.

\textit{Memory read.}
A frozen DINOv3 encoder $\Phi(\cdot)$ is used to construct retrieval features independently of the SAM2 segmentation backbone. For an input image $x_t$, the normalized query feature is
\begin{equation}
q_t =
\frac{\Phi(x_t)}
{\|\Phi(x_t)\|_2}.
\label{eq:dino_query}
\end{equation}
Each few-shot or working-memory entry is associated with a DINOv3 retrieval key and a SAM2-compatible memory representation. The controller ranks available entries by cosine similarity to $q_t$ and retrieves the top-$k$ relevant memories:
\begin{equation}
\mathcal{M}_{t}^{\mathrm{read}}
=
\mathcal{M}^{s}
\cup
\operatorname{TopK}
\left(
q_t,
\mathcal{M}^{f}\cup\mathcal{M}^{w}_{t-1}
\right).
\label{eq:controller_read}
\end{equation}
The supervised static memory $\mathcal{M}^{s}$ is retained as a persistent anchor, while few-shot and working memories provide query-dependent information when available. DINOv3 is used only for memory selection; the retrieved values remain SAM2-compatible representations and are composed through the pretrained SAM2 memory-attention module.

\textit{Memory write.}
When a clinician provides a corrected annotation $y_t^{*}$, the corrected observation is encoded through the same frozen SAM2 observation-to-memory pathway used for few-shot examples:
\begin{equation}
m_t^{c}
=
\mathcal{E}_{m}
\left(
\mathcal{E}_{I}(x_t),
y_t^{*}
\right),
\label{eq:corrected_memory}
\end{equation}
with the corresponding normalized retrieval key
\begin{equation}
k_t^{c}
=
\frac{\Phi(x_t)}
{\|\Phi(x_t)\|_2}.
\label{eq:corrected_key}
\end{equation}
This shared encoding pathway ensures that few-shot and clinician-corrected observations remain compatible with the native SAM2 memory representation.

For a non-empty working-memory buffer, the controller identifies the closest existing entry according to
\begin{equation}
j^{*}
=
\arg\max_j
\operatorname{sim}
\left(
k_t^{c},k_j^{w}
\right),
\qquad
\rho_t
=
\max_j
\operatorname{sim}
\left(
k_t^{c},k_j^{w}
\right).
\label{eq:write_similarity}
\end{equation}
If the buffer is empty or $\rho_t<\tau_{\mathrm{sim}}$, the corrected observation is treated as a novel target-domain pattern and appended to the working-memory bank. Otherwise, the correction is consolidated with its closest entry using a delta update:
\begin{equation}
\Delta m_t
=
m_t^{c}-m_{j^{*}}^{w},
\qquad
m_{j^{*}}^{w}
\leftarrow
m_{j^{*}}^{w}
+
\eta\Delta m_t,
\label{eq:delta_update}
\end{equation}
where $\eta\in(0,1]$ controls the consolidation strength. The associated retrieval key is updated consistently:
\begin{equation}
k_{j^{*}}^{w}
\leftarrow
\operatorname{Norm}
\left[
(1-\eta)k_{j^{*}}^{w}
+
\eta k_t^{c}
\right].
\label{eq:key_update}
\end{equation}
The write policy can therefore be summarized as
\begin{equation}
\begin{aligned}
\mathcal{A}_t &=
\operatorname{Append}
\left(
\mathcal{M}^{w}_{t-1},
(m_t^{c},k_t^{c})
\right),\\
\mathcal{C}_t &=
\operatorname{Consolidate}
\left(
\mathcal{M}^{w}_{t-1},
j^{*},m_t^{c},k_t^{c}
\right).
\end{aligned}
\label{eq:memory_operations}
\end{equation}

\begin{equation}
\mathcal{M}^{w}_{t}
=
\begin{cases}
\mathcal{A}_t,
&
\mathcal{M}^{w}_{t-1}=\varnothing
\ \text{or}\ 
\rho_t<\tau_{\mathrm{sim}},\\[3pt]
\mathcal{C}_t,
&
\text{otherwise}.
\end{cases}
\label{eq:controller_write}
\end{equation}

Similarity-aware consolidation keeps working memory compact by merging updates into sufficiently similar existing entries, while genuinely novel observations are still appended as new states, preventing redundant growth without discarding anything new. Consequently, the controller separates stable source knowledge from continuously evolving deployment experience: during reading operation, it preserves the static-memory anchor while selectively retrieving and integrating relevant target-domain information for prediction; and during writing operation, it determines whether newly acquired clinician feedback should expand the working-memory buffer or be consolidated into an existing persistent entry.

\section{Experiments}
\label{sec:experiments}

\subsection{Datasets and Preprocessing}

We evaluate the proposed framework on four public datasets covering abdominal MRI, cardiac MRI, and echocardiography. \textbf{CHAOS}~\cite{kavur2021chaos} provides abdominal MRI with left kidney(LK), right kidney(RK), spleen(Spl.), and liver(Liv.) annotations. \textbf{ACDC}~\cite{bernard2018deep} contains cardiac MRI with right ventricle (RV), myocardium (Myo), and left ventricle (LV) annotations at end-diastole (ED) and end-systole (ES). \textbf{CAMUS}~\cite{leclerc2019deep} consists of two- and four-chamber echocardiographic views with LV, Myo, and left atrium (LA) annotations. For cross-domain evaluation, we use \textbf{CardiacUDA}~\cite{yang2023cardiacUDA} as an external target domain. Specifically, we use the public \texttt{labe\_all\_frame} subset, which provides dense frame-wise annotations suitable for full-sequence evaluation, whereas the remaining released subsets are sparsely annotated or, in some cases, unlabeled. We exclude volumes containing label codes outside the documented A4C label space, leaving eight valid full-frame patient volumes comprising 428 slices in total. Thus, the resulting cohort includes all usable volumes from the densely annotated subset under our preprocessing criteria rather than a manually selected sample. Because the public release does not provide a case-level manifest corresponding to its reported 8:1:1 split, we treat this cohort as a fixed external CardiacUDA subset rather than the complete official test partition. CAMUS and CardiacUDA share LV and LA labels, enabling direct evaluation of cross-domain generalization and test-time adaptation.
\begin{table*}[!t]
\centering
\caption{Single-class segmentation under different supervision levels on CHAOS, ACDC, and CAMUS. Dice (\%) and HD95 (mm).}
\label{tab:static_results}
\setlength{\tabcolsep}{3.5pt}
\renewcommand{\arraystretch}{1.15}

\resizebox{\textwidth}{!}{%
\begin{tabular}{c l cc cc cc cc cc cc cc cc cc cc}
\toprule

& & \multicolumn{8}{c}{CHAOS~\cite{CHAOSdata2019}}
& \multicolumn{6}{c}{ACDC~\cite{bernard2018deep}}
& \multicolumn{6}{c}{CAMUS~\cite{CAMUS}} \\
\cmidrule(lr){3-10}
\cmidrule(lr){11-16}
\cmidrule(lr){17-22}

& & \multicolumn{2}{c}{LK}
& \multicolumn{2}{c}{RK}
& \multicolumn{2}{c}{Spl.}
& \multicolumn{2}{c}{Liv.}
& \multicolumn{2}{c}{RV}
& \multicolumn{2}{c}{Myo}
& \multicolumn{2}{c}{LV}
& \multicolumn{2}{c}{LV}
& \multicolumn{2}{c}{Myo}
& \multicolumn{2}{c}{LA} \\

\cmidrule(lr){3-4}
\cmidrule(lr){5-6}
\cmidrule(lr){7-8}
\cmidrule(lr){9-10}
\cmidrule(lr){11-12}
\cmidrule(lr){13-14}
\cmidrule(lr){15-16}
\cmidrule(lr){17-18}
\cmidrule(lr){19-20}
\cmidrule(lr){21-22}

Sup. & Method
& Dice$\uparrow$ & HD95$\downarrow$
& Dice$\uparrow$ & HD95$\downarrow$
& Dice$\uparrow$ & HD95$\downarrow$
& Dice$\uparrow$ & HD95$\downarrow$
& Dice$\uparrow$ & HD95$\downarrow$
& Dice$\uparrow$ & HD95$\downarrow$
& Dice$\uparrow$ & HD95$\downarrow$
& Dice$\uparrow$ & HD95$\downarrow$
& Dice$\uparrow$ & HD95$\downarrow$
& Dice$\uparrow$ & HD95$\downarrow$ \\
\midrule

\multirow{7}{*}{100\%}
& U-Net~\cite{ronneberger2015unet}
& 76.67 & 43.42
& 71.37 & 90.12
& 67.91 & 35.03
& 87.35 & 13.37
& 82.03 & 6.54
& 86.83 & 3.05
& \underline{92.18} & 4.28
& 93.36 & 3.03
& 86.97 & 5.00
& 90.48 & 4.53 \\

& SwinUNETR~\cite{hatamizadeh2022swin}
& 75.34 & 94.75
& 80.98 & 124.78
& 72.32 & 111.60
& 83.15 & 54.21
& 71.06 & 49.87
& 79.67 & 51.16
& 80.69 & 46.57
& 93.32 & 2.91
& \textbf{94.76} & \textbf{1.62}
& \textbf{96.22} & \textbf{1.64} \\

& nnU-Net~\cite{isensee2021nnunet}
& 89.15 & 14.57
& 84.91 & 15.73
& 82.82 & 11.36
& 89.26 & 10.76
& \textbf{90.53} & \textbf{1.62}
& \textbf{90.36} & \textbf{1.12}
& \textbf{94.94} & \textbf{1.20}
& \textbf{94.61} & \textbf{1.97}
& \underline{89.94} & \underline{3.01}
& \underline{92.67} & \underline{3.01} \\

& MedSAM2(Full FT)~\cite{ma2025medsam2}
& 85.42 & 23.05
& 87.49 & 18.90
& 77.45 & 37.08
& 88.65 & 15.62
& \underline{87.89} & 6.27
& \underline{88.03} & \underline{1.37}
& 92.05 & 4.13
& \underline{93.62} & \underline{2.34}
& 87.75 & 4.05
& 90.39 & 4.59 \\

& \textbf{Ours - Static Memory (MedSAM2)}
& 87.02 & 4.41
& 84.91 & 4.71
& 82.03 & 9.74
& 82.03 & 7.86
& 80.09 & 3.70
& 79.75 & 2.83
& 89.29 & 2.88
& 91.05 & 4.30
& 81.37 & 7.68
& 86.89 & 6.34 \\

& \textbf{Ours -- Static Memory (SAM2)}
& \textbf{92.99} & \underline{3.76}
& \textbf{94.10} & \underline{2.32}
& \underline{88.59} & \textbf{1.95}
& \underline{90.21} & \underline{3.40}
& 85.26 & \underline{2.83}
& 84.74 & 1.78
& 91.75 & 2.00
& 92.66 & 3.05
& 85.75 & 4.96
& 90.12 & 4.27 \\

& \textbf{Ours -- Static Memory (SAM2) -- FL}
& \underline{92.08} & \textbf{3.20}
& \underline{93.14} & \textbf{1.25}
& \textbf{90.27} & \underline{2.60}
& \textbf{91.71} & \textbf{2.93}
& 79.54 & 2.98
& 83.03 & 1.70
& 91.39 & \underline{1.60}
& 92.34 & 3.14
& 85.43 & 4.97
& 89.67 & 4.88 \\
\midrule

\multirow{6}{*}{30\%}
& U-Net~\cite{ronneberger2015unet}
& 58.33 & 58.94
& 61.06 & 100.93
& 53.10 & 59.73
& 74.10 & 26.05
& 75.80 & 15.21
& 82.33 & 4.97
& 89.63 & 9.33
& 92.49 & 4.03
& 85.57 & 6.34
& 88.98 & 6.43 \\

& SwinUNETR~\cite{hatamizadeh2022swin}
& 44.62 & 129.70
& 56.90 & 138.12
& 35.50 & 139.67
& 65.13 & 59.24
& 49.53 & 71.14
& 54.90 & 70.96
& 69.17 & 83.78
& 92.08 & 4.23
& 85.11 & 6.23
& \underline{89.38} & 6.01 \\

& nnU-Net~\cite{isensee2021nnunet}
& \textbf{87.69} & 7.58
& \underline{86.20} & \underline{5.72}
& \textbf{81.50} & 17.23
& \textbf{87.41} & \textbf{5.59}
& \textbf{87.75} & \textbf{2.87}
& \textbf{87.32} & 2.80
& \textbf{93.37} & \textbf{1.60}
& \textbf{93.64} & \textbf{2.52}
& \textbf{87.80} & \textbf{3.99}
& \textbf{91.61} & \textbf{3.96} \\

& MedSAM2(Full FT)~\cite{ma2025medsam2}
& 44.30 & 103.71
& 78.16 & 42.02
& \underline{69.15} & 51.21
& 82.73 & 28.37
& \underline{83.06} & 11.48
& \underline{86.17} & \textbf{1.98}
& \underline{91.78} & 4.12
& \underline{92.86} & \underline{3.01}
& \underline{86.29} & \underline{5.32}
& 89.15 & \underline{4.48} \\

& \textbf{Ours - Static Memory (MedSAM2)}
& 64.30 & \underline{7.41}
& 80.75 & 6.60
& 61.59 & \underline{16.19}
& 75.71 & 9.11
& 75.46 & 4.62
& 77.90 & 3.26
& 88.11 & 3.25
& 90.05 & 5.21
& 80.04 & 8.61
& 85.12 & 7.72 \\

& \textbf{Ours -- Static Memory (SAM2)}
& \underline{84.32} & \textbf{5.32}
& \textbf{86.63} & \textbf{2.74}
& 65.78 & \textbf{6.56}
& \underline{84.55} & \underline{8.37}
& 78.60 & \underline{3.27}
& 81.79 & \underline{2.35}
& 90.70 & \underline{2.35}
& 91.63 & 3.85
& 83.65 & 6.33
& 87.39 & 6.03 \\
\midrule

\multirow{6}{*}{10\%}
& U-Net~\cite{ronneberger2015unet}
& 47.59 & 88.74
& 50.74 & 102.75
& 21.30 & 107.76
& 37.83 & 63.64
& 43.20 & 74.20
& 64.59 & 35.93
& 74.51 & 39.44
& 90.87 & 6.40
& 83.02 & 8.86
& 85.98 & 11.78 \\

& SwinUNETR~\cite{hatamizadeh2022swin}
& 33.07 & 145.09
& 31.54 & 157.03
& 12.15 & 134.64
& 38.79 & 69.20
& 50.04 & 75.02
& 55.01 & 73.91
& 70.40 & 82.64
& 90.51 & 6.25
& 81.93 & 9.55
& 86.73 & 9.89 \\

& nnU-Net~\cite{isensee2021nnunet}
& \textbf{76.83} & 32.82
& \underline{78.19} & 36.02
& \textbf{68.35} & \underline{48.09}
& \textbf{84.20} & 28.09
& \textbf{81.72} & 6.64
& \textbf{83.41} & \textbf{2.60}
& \textbf{90.19} & \textbf{3.75}
& \textbf{92.88} & \textbf{3.25}
& \textbf{86.23} & \textbf{4.93}
& \textbf{90.45} & \textbf{5.30} \\

& MedSAM2(Full FT)~\cite{ma2025medsam2}
& 58.96 & 76.46
& 70.25 & 59.99
& \underline{62.21} & 81.03
& \underline{78.53} & 38.85
& \underline{79.81} & 13.54
& \underline{81.61} & \underline{3.41}
& \underline{88.76} & 6.24
& \underline{91.85} & \underline{4.12}
& \underline{83.46} & 8.75
& \underline{87.57} & 7.23 \\

& \textbf{Ours - Static Memory (MedSAM2)}
& 49.92 & \textbf{9.80}
& 52.90 & \underline{8.44}
& 10.61 & 75.65
& 40.58 & \underline{21.42}
& 61.61 & \underline{6.51}
& 73.33 & 4.10
& 85.23 & 4.78
& 88.69 & 6.11
& 77.58 & 10.50
& 83.08 & 9.14 \\

& \textbf{Ours -- Static Memory (SAM2)}
& \underline{66.29} & \underline{10.69}
& \textbf{80.10} & \textbf{5.21}
& 56.18 & \textbf{11.17}
& 65.34 & \textbf{15.05}
& 72.71 & \textbf{6.30}
& 78.00 & 3.72
& 87.92 & \underline{5.50}
& 90.38 & 4.81
& 81.30 & \underline{8.24}
& 85.99 & \underline{7.00} \\

\bottomrule
\end{tabular}%
}
\end{table*}

All experiments are formulated as class-specific binary segmentation tasks. Images are resampled to an in-plane resolution of $0.5\times0.5$~mm and center-cropped to $256\times256$~pixels. For CHAOS, ACDC, and CAMUS,  we use all 40, 100, and 500 patients, respectively. We perform patient-wise five-fold cross-validation and data are split at the patient level into training, validation, and test sets with a ratio of 3:1:1. Segmentation results are averaged across the five test folds. CardiacUDA serves solely as the fixed external evaluation cohort and is excluded from training or validation.

\textbf{Ethics statement.}
This study used only publicly available, de-identified datasets, including CHAOS, ACDC, CAMUS, and CardiacUDA \cite{kavur2021chaos,bernard2018deep,leclerc2019deep,yang2023cardiacUDA}. No identifiable patient information was accessed and no new human participants were recruited. Accordingly, no additional institutional review board approval or informed consent was required for this secondary analysis.

\subsection{Implementation Details}

All experiments were implemented in PyTorch and conducted on an NVIDIA Blackwell PRO 6000 GPU. U-Net~\cite{ronneberger2015unet}, SwinUNETR~\cite{hatamizadeh2022swin}, MedSAM2~\cite{ma2025medsam2}, and the supervised static-memory component of MemSeg-Agent were trained for 200 epochs with a batch size of 8. nnU-Net~\cite{isensee2021nnunet} followed its official configuration and was trained for 1000 epochs with a batch size of 2. AdamW was used with a learning rate of $3\times10^{-3}$ and weight decay of $10^{-4}$.

MemSeg-Agent uses SAM2.1 Hiera-Tiny as the segmentation backbone. Its image encoder, memory encoder, memory attention, prompt encoder, and mask decoder remain frozen throughout training and deployment, such that only the static-memory variables are optimized during supervised memory construction. Few-shot and test-time working memories require no gradient-based optimization. Prediction uses an empty prompt, and a frozen DINOv3 encoder is used for similarity-based memory retrieval. Training and inference are performed slice-wise, while segmentation performance is evaluated after reconstructing three-dimensional volumes using the Dice similarity coefficient (Dice) for region overlap and the 95th-percentile Hausdorff distance (HD95) for boundary accuracy.

For clinician-triggered adaptation, target volumes are processed sequentially under the strict prequential protocol described in Section~\ref{sec:working_memory}. In retrospective experiments, clinician dissatisfaction is simulated using $\tau_{\mathrm{corr}}=0.5$, while working-memory novelty is determined using $\tau_{\mathrm{sim}}=0.95$. 

\subsection{Cross-Domain Working-Memory Adaptation}

In clinical deployment, models inevitably encounter domain shift when transferred to new scanners or imaging protocols. Continual segmentation methods address this through sequential parameter updates. We evaluate whether working-memory adaptation can offer an effective solution. We compare MemSeg-Agent with the official Lifelong nnU-Net~\cite{gonzalez2023lifelongnnunet} and UnCLeSAM~\cite{ranem2024unclesam}. We evaluate this setting by transferring models trained on CAMUS to the external CardiacUDA cohort and compare MemSeg-Agent with Lifelong nnU-Net~\cite{gonzalez2023lifelongnnunet} and UnCLeSAM~\cite{ranem2024unclesam}. All methods are evaluated on the same target volumes using identical volume-level metrics. As shown in Table~\ref{tab:domain_shift}, direct transfer yields macro-average Dice scores of 19.79\%, 19.90\%, and 15.57\% for static-only MemSeg-Agent, Lifelong nnU-Net, and UnCLeSAM, respectively. With working-memory adaptation, MemSeg-Agent achieves 77.93\% Dice and 9.90~mm HD95 on CardiacUDA, compared with 19.79\%/44.02~mm for static memory alone, 19.90\%/86.63~mm for Lifelong nnU-Net, and 15.57\%/97.59~mm for UnCLeSAM.

\begin{table}[h]
\centering
\caption{Source-domain performance and cross-domain degradation.
Values are macro-averages over LV and LA. $\Delta$ denotes
CardiacUDA minus CAMUS.}
\label{tab:domain_shift}
\setlength{\tabcolsep}{4pt}
\resizebox{\columnwidth}{!}{%
\begin{tabular}{lccc}
\toprule
\multicolumn{4}{c}{Dice (\%) $\uparrow$} \\
\midrule
Method 
& \makecell[c]{CAMUS \\ \small\cite{CAMUS}} 
& \makecell[c]{CardiacUDA \\ \small\cite{yang2023cardiacUDA}} 
& $\Delta$ \\
\midrule
\makecell[l]{UnCLeSAM \\ \small\cite{ranem2024unclesam}}
    & 82.21 & 15.57 & -66.63 \\[3pt]
\makecell[l]{Lifelong nnU-Net \\ \small\cite{gonzalez2023lifelongnnunet}}
    & \textbf{92.64} & 19.90 & -72.74 \\[3pt]
MemSeg-Agent (Static-only)
    & 90.45 & 19.79 & -70.66 \\
MemSeg-Agent
    & 90.45 & \textbf{77.93} & \textbf{-12.52} \\
\bottomrule
\end{tabular}%
}
\end{table}

\begin{figure*}[!t]
\centering
\includegraphics[width=\textwidth]{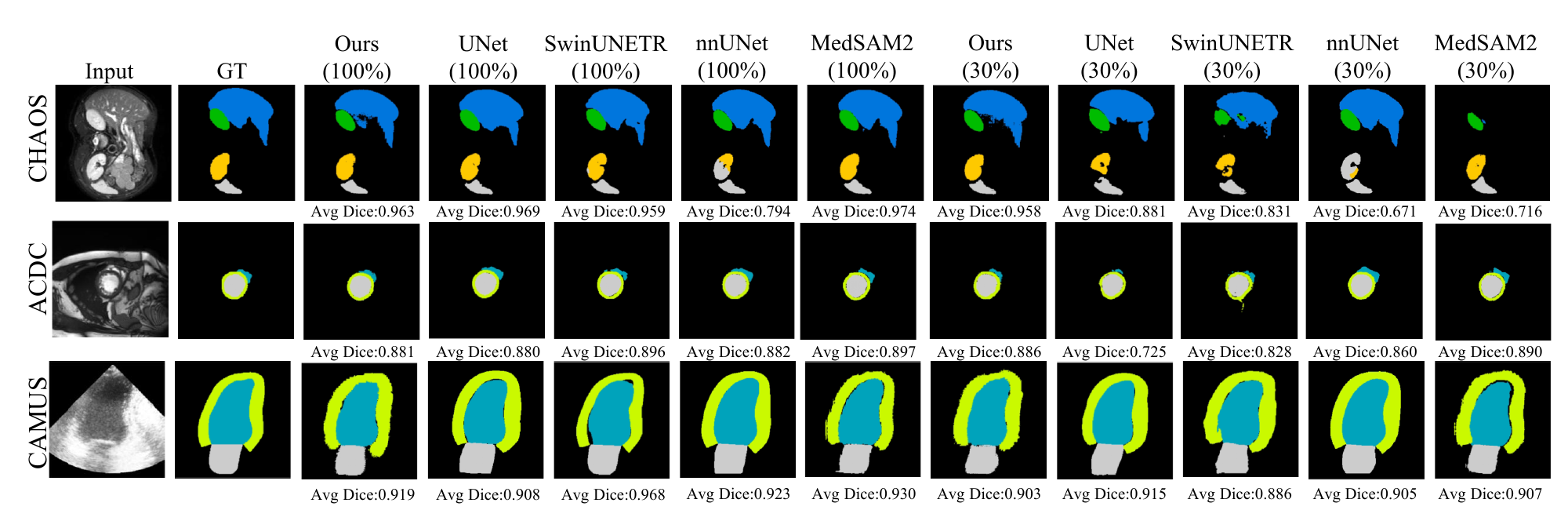}
\caption{
Qualitative segmentation results on CHAOS, ACDC, and CAMUS under 100\% and 30\% supervision. 
Average Dice of all classes scores are shown below each prediction. }
\label{fig:visual_comparison}
\end{figure*}





\subsection{Static Memory via Supervised and Federated Learning}

Conventional segmentation methods acquire task-specific knowledge by fine-tuning model parameters on annotated data. We evaluate whether static memory can serve as an equivalent substrate under varying annotation budgets (10\%, 30\%, 100\% supervision) across three datasets, and whether memory-only exchange enables efficient federated collaboration across institutions. We demonstrate the performance of static memory in Table~\ref{tab:static_results} and qualitative results in Fig.~\ref{fig:visual_comparison}. We evaluate the
same pseudo-observation memory parameterization with two frozen
SAM2-family backbones: the default SAM2.1 Hiera-Tiny backbone and the
official MedSAM2 backbone, respectively. Static memory remains competitive with fully trained segmentation models while keeping the backbone frozen.

The memory formulation also supports federated learning settings. In the four-site FedAvg experiment, only the pseudo-observation memory is exchanged across sites. Since only a 2M-parameter memory is updated rather than the full 148.63M SAM2-tiny backbone, per-round communication is reduced by $\sim$74.3$\times$ ($\sim$98.65\%) while maintaining competitive segmentation performance.

\subsection{Working-Memory Accumulation and Consolidation}



We further evaluate how working memory accumulates deployment experience and whether similarity-aware consolidation can control memory growth during sequential adaptation from CAMUS to CardiacUDA. As shown in Table~\ref{tab:cross_domain}, static memory alone achieves 31.16\%/8.42\% Dice for LV/LA. Static+Append improves performance to 83.10\%/73.38\%, while Static+Delta achieves comparable performance of 82.41\%/73.45\%. On the final five target volumes, Static+Delta achieves 81.12\%/69.73\% Dice, compared with 33.06\%/4.79\% for static memory alone. Meanwhile, Delta consolidation reduces the final working-memory size from 29/114 entries with append-only storage to 22/40 entries for LV/LA. We additionally visualize sequential adaptation under the CHAOS$\rightarrow$ACDC shift in Fig.~\ref{fig:test-time-adaptation}, where segmentation performance improves as target-domain corrections accumulate in working memory.

\begin{table}[!t]
\centering
\caption{Cross-domain adaptation from CAMUS to CardiacUDA. Dice (\%) and HD95 (mm) are reported for the full stream; Subseq. Dice evaluates the final five volumes.}
\label{tab:cross_domain}
\setlength{\tabcolsep}{3.0pt}
\renewcommand{\arraystretch}{1.10}
\resizebox{\columnwidth}{!}{%
\begin{tabular}{lccccccc}
\toprule
& \multicolumn{2}{c}{LV}
& \multicolumn{2}{c}{LA}
& LV
& LA
& WM \\
\cmidrule(lr){2-3}
\cmidrule(lr){4-5}
\cmidrule(lr){6-7}
Method
& Dice$\uparrow$ & HD95$\downarrow$
& Dice$\uparrow$ & HD95$\downarrow$
& \multicolumn{2}{c}{Subseq. Dice$\uparrow$}
& Final entries$\downarrow$ \\
\midrule
Static only
& 31.16 & 47.91
& 8.42 & 40.13
& 33.06 & 4.79
& 0/0 \\

Static + Append
& \textbf{83.10} & \textbf{11.54}
& 73.38 & \textbf{6.86}
& \textbf{82.22} & 69.30
& 29/114 \\

Static + Delta
& 82.41 & 11.74
& \textbf{73.45} & 8.07
& 81.12 & \textbf{69.73}
& \textbf{22/40} \\
\bottomrule
\end{tabular}%
}
\end{table}

\begin{figure*}[!t]
\centering

\begin{minipage}[t]{0.49\textwidth}
    \centering
    \includegraphics[width=\linewidth]{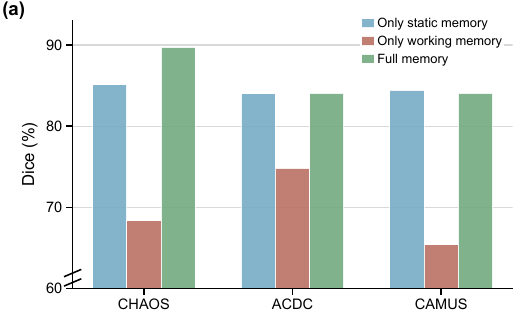}
\end{minipage}\hfill
\begin{minipage}[t]{0.49\textwidth}
    \centering
    \includegraphics[width=\linewidth]{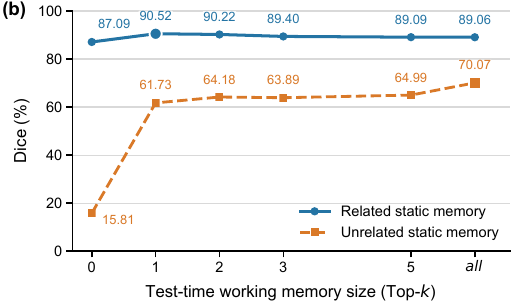}
\end{minipage}

\caption{Complementary roles of static and working memory.
(a) Dice comparison of only static memory, only working memory, and full memory across three datasets.
(b) Sensitivity to test-time working-memory size under related and unrelated static memories. Related denotes CHAOS-trained static memory evaluated on CHAOS, while unrelated denotes CHAOS-trained static memory evaluated on ACDC.}
\label{fig:memory_ablation}
\end{figure*}

\begin{center}
\includegraphics[width=\linewidth]{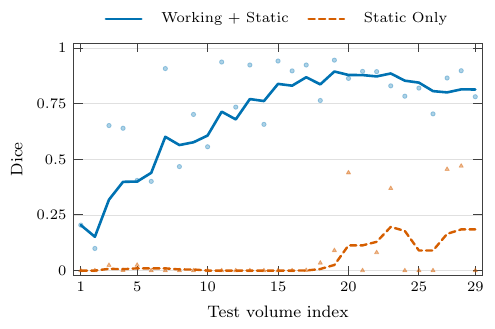}
\captionof{figure}{Sequential test-time adaptation under the CHAOS$\rightarrow$ACDC
    domain shift. Markers denote per-volume Dice scores, whereas curves show
    the corresponding 5-volume trailing moving averages.}
\label{fig:test-time-adaptation}
\end{center}

\subsection{Ablation Study}


We first examine whether segmentation performance depends on meaningful memory content. As shown in Fig.~\ref{fig:memory-validity}, task-aligned memories, including pseudo-observations, direct memory tokens, and real examples, produce meaningful predictions, whereas unrelated-domain, label-shuffled, random, and empty memories yield near-zero Dice ($\leq 0.26\%$). Among the evaluated representations, direct memory tokens achieve the highest Dice, while pseudo-observations remain effective and preserve the shared SAM2-compatible memory interface used throughout MemSeg-Agent.

Figure~\ref{fig:memory_ablation}(a) compares static-only, working-only, and combined memory, with the combined setting achieving the strongest overall performance across all evaluated datasets. Figure~\ref{fig:memory_ablation}(b) shows that fewer working-memory entries are sufficient when a related static prior is available, while performance benefits more from additional working memory when the static prior is unrelated.

The same static-memory formulation also remains effective with a frozen MedSAM2 backbone, as reported in Table~\ref{tab:static_results}, further supporting its applicability across different SAM2-family backbones.

\begin{center}
\includegraphics[width=\linewidth]{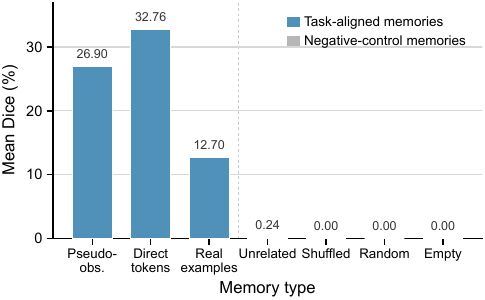}
\captionof{figure}{Memory-validity analysis on CardiacUDA using mean volume-level Dice (\%) across LV and LA. Task-aligned memories yield meaningful predictions, while negative-control memories result in near-zero Dice results. Pseudo-obs. denotes the pseudo-observation representation adopted throughout MemSeg-Agent.}
\label{fig:memory-validity}
\end{center}

\section{Discussion}
\label{sec:discussion}
Our results demonstrate that task- and domain-specific knowledge can be effectively represented and updated in the native memory space of a frozen segmentation foundation model. Under cross-domain deployment, all evaluated methods suffer substantial performance degradation under direct transfer from CAMUS to CardiacUDA; in contrast, incorporating target-domain corrections through working memory markedly improves subsequent segmentation without modifying the pretrained backbone, demonstrating that target-domain experience can be effectively accumulated in memory space. In supervised settings, static memory remains competitive across imaging modalities, anatomical regions, and annotation budgets, indicating that task-specific knowledge can be encoded outside the backbone parameters. And memory-only federated learning further extends this finding by substantially reducing per-round communication—approximately 74.3$\times$—through the exchange of only compact memory variables. During sequential deployment, delta-based consolidation achieves performance comparable to append-only storage with a smaller working-memory bank, indicating that redundant deployment experience can be compressed without loss. Finally, the ablation studies confirm that performance depends on task-relevant memory content, that static and working memories provide complementary sources of prior knowledge and deployment experience, and that the same static-memory formulation remains effective across SAM2-family backbones. Together, these findings support memory space as a shared substrate for knowledge acquisition, deployment-time adaptation, and reuse across different stages of the model lifecycle.

Several limitations remain. First, clinician feedback is currently simulated retrospectively using ground-truth corrections. Real-world clinical validation has not yet been conducted. Therefore, the current results may not fully represent performance under realistic clinical feedback. variability, or clinical correction time. Second, the current experiments use class-specific, slice-wise binary segmentation; extension to joint multi-class and three-dimensional segmentation requires further validation. Third, static-memory performance degrades under very limited supervision, partly due to only four patient volumes are available for static-memory training under the 10\% supervision setting, which substantially limits the diversity and reliability of the learned task prior.  Fourth, the controller is also lightweight and rule-based, and its retrieval thresholds and consolidation criteria may require adjustment across different deployment domains. Finally, the federated experiment demonstrates communication efficiency but does not provide formal privacy guarantees. Future work will evaluate the framework on larger multi-institutional cohorts with realistic clinician feedback, and extend memory-space adaptation to multi-class and volumetric segmentation with more adaptive retrieval and consolidation policies.

Beyond medical segmentation, MemSeg-Agent suggests a different interpretation of test-time learning. Existing test-time training (TTT) and adaptation methods typically introduce fast plasticity by optimizing selected parameters, auxiliary modules, or other fast weights during deployment, while long-term knowledge remains encoded in slowly learned model parameters. Biological learning, however, does not exhibit a strict separation between training and inference. Instead, information is integrated over multiple timescales, with rapidly changing neural activity and transient memory traces supporting immediate behavior, while slower consolidation stabilizes selected experience into longer-term knowledge. MemSeg-Agent conceptually follows this multi-timescale organization: the frozen foundation model provides a stable long-term prior, working memory serves as a rapidly editable substrate for incorporating sparse target-domain experience, and memory consolidation determines which information persists and influences subsequent cases. From this perspective, the central idea is not merely to replace parameter updates with a memory controller, but to move from weight-centric adaptation toward memory-centric learning, where fast adaptation, selective consolidation, and stable prior knowledge are represented as distinct but interacting mechanisms. This formulation may provide a broader direction for continual and test-time learning systems that remain plastic in changing environments, where adaptation knowledge accumulates in memory rather than overwrites the model itself.

\section{Conclusion}
\label{sec:conclusion}

We introduced \emph{memory-space adaptation} as a unified learning formulation for medical image segmentation, in which task knowledge and deployment experience are maintained through an explicit memory state while the foundation model remains fixed. MemSeg-Agent realizes this formulation through complementary static, few-shot, and working memories coordinated by a shared read--write process. Across supervised, continual, cross-domain, and federated settings, our results show that adaptation knowledge can be stored, accumulated, and reused effectively in memory space, supporting a more general view of learning that extends across different supervision levels and deployment stages. Future work will explore this paradigm in multi-class volumetric segmentation, larger multi-institutional cohorts, and prospective clinical settings with realistic and potentially noisy clinician feedback.

\section*{Acknowledgment}
The authors gratefully acknowledge support from the NVIDIA Academic Grant Program for providing computational resources used in this work.

\section*{Declaration of generative AI and AI-assisted technologies in the manuscript preparation process}

During the preparation of this work, the authors used OpenAI's ChatGPT solely to improve the language and readability of the manuscript. The authors reviewed and edited the AI-assisted text and take full responsibility for the content of the publication.

\bibliographystyle{cas-model2-names}
\bibliography{references}

@article{azad2024medical,
  title={Medical image segmentation review: The success of u-net},
  author={Azad, Reza and Aghdam, Ehsan Khodapanah and Rauland, Amelie and Jia, Yiwei and Avval, Atlas Haddadi and Bozorgpour, Afshin and Karimijafarbigloo, Sanaz and Cohen, Joseph Paul and Adeli, Ehsan and Merhof, Dorit},
  journal={IEEE Transactions on Pattern Analysis and Machine Intelligence},
  year={2024},
  publisher={IEEE}
}

@article{kirillov2023segment,
  title={Segment Anything},
  author={Kirillov, Alexander and others},
  journal={Proceedings of the IEEE/CVF International Conference on Computer Vision},
  year={2023}
}

@article{ma2024medsam,
  title={MedSAM: Segment Anything in Medical Images},
  author={Ma, Jun and others},
  journal={arXiv preprint arXiv:2304.12306},
  year={2024}
}

@article{sheller2020federated,
  title={Federated learning in medicine: facilitating multi-institutional collaborations without sharing patient data},
  author={Sheller, Micah J. and others},
  journal={Scientific Reports},
  volume={10},
  number={1},
  pages={12598},
  year={2020}
}

@article{litjens2017survey,
  title={A survey on deep learning in medical image analysis},
  author={Litjens, Geert and others},
  journal={Medical Image Analysis},
  volume={42},
  pages={60--88},
  year={2017}
}

@inproceedings{ronneberger2015unet,
  title={U-Net: Convolutional networks for biomedical image segmentation},
  author={Ronneberger, Olaf and Fischer, Philipp and Brox, Thomas},
  booktitle={MICCAI},
  year={2015}
}

@article{kelly2019key,
  title={Key challenges for delivering clinical impact with artificial intelligence},
  author={Kelly, Christopher J. and others},
  journal={BMC Medicine},
  volume={17},
  number={1},
  pages={195},
  year={2019}
}

@article{ravi2024sam,
  title={Sam 2: Segment anything in images and videos},
  author={Ravi, Nikhila and Gabeur, Valentin and Hu, Yuan-Ting and Hu, Ronghang and Ryali, Chaitanya and Ma, Tengyu and Khedr, Haitham and R{\"a}dle, Roman and Rolland, Chloe and Gustafson, Laura and others},
  journal={arXiv preprint arXiv:2408.00714},
  year={2024}
}

@article{oquab2023dinov2,
  title={Dinov2: Learning robust visual features without supervision},
  author={Oquab, Maxime and Darcet, Timoth{\'e}e and Moutakanni, Th{\'e}o and Vo, Huy and Szafraniec, Marc and Khalidov, Vasil and Fernandez, Pierre and Haziza, Daniel and Massa, Francisco and El-Nouby, Alaaeldin and others},
  journal={arXiv preprint arXiv:2304.07193},
  year={2023}
}

@article{zhao2025retrieval,
  title={Retrieval-augmented few-shot medical image segmentation with foundation models},
  author={Zhao, Lin and Chen, Xiao and Chen, Eric Z and Liu, Yikang and Chen, Terrence and Sun, Shanhui},
  journal={IEEE Transactions on Neural Networks and Learning Systems},
  year={2025},
  publisher={IEEE}
}

@inproceedings{zhang2025adapting,
  title={Adapting vision foundation models for real-time ultrasound image segmentation},
  author={Zhang, Xiaoran and Chen, Eric Z and Zhao, Lin and Chen, Xiao and Liu, Yikang and Maihe, Boris and Duncan, James S and Chen, Terrence and Sun, Shanhui},
  booktitle={International Conference on Medical Image Computing and Computer-Assisted Intervention},
  pages={24--34},
  year={2025},
  organization={Springer}
}

@article{ma2025medsam2,
  title={Medsam2: Segment anything in 3d medical images and videos},
  author={Ma, Jun and Yang, Zongxin and Kim, Sumin and Chen, Bihui and Baharoon, Mohammed and Fallahpour, Adibvafa and Asakereh, Reza and Lyu, Hongwei and Wang, Bo},
  journal={arXiv preprint arXiv:2504.03600},
  year={2025}
}

@article{ma2024segment,
  title={Segment anything in medical images},
  author={Ma, Jun and He, Yuting and Li, Feifei and Han, Lin and You, Chenyu and Wang, Bo},
  journal={Nature communications},
  volume={15},
  number={1},
  pages={654},
  year={2024},
  publisher={Nature Publishing Group UK London}
}

@article{fallahpour2025medrax,
  title={Medrax: Medical reasoning agent for chest x-ray},
  author={Fallahpour, Adibvafa and Ma, Jun and Munim, Alif and Lyu, Hongwei and Wang, Bo},
  journal={arXiv preprint arXiv:2502.02673},
  year={2025}
}

@article{jia2026agentic,
  title={Agentic memory-augmented retrieval and evidence grounding for medical question-answering tasks},
  author={Jia, Shuyue and Bit, Subhrangshu and Jasodanand, Varuna H and Liu, Yi and Kolachalama, Vijaya B},
  journal={International Journal of Medical Informatics},
  pages={106339},
  year={2026},
  publisher={Elsevier}
}

@article{isensee2021nnunet,
  title={nnU-Net: a self-configuring method for deep learning-based biomedical image segmentation},
  author={Isensee, Fabian and Jaeger, Paul F. and Kohl, Simon A. A. and Petersen, Jens and Maier-Hein, Klaus H.},
  journal={Nature Methods},
  volume={18},
  pages={203--211},
  year={2021},
  publisher={Nature Publishing Group},
  doi={10.1038/s41592-020-01008-z}
}

@article{hatamizadeh2022swin,
  title={Swin UNETR: Swin Transformers for Semantic Segmentation of Brain Tumors in MRI Images},
  author={Hatamizadeh, Ali and Nath, Vishwesh and Tang, Yucheng and Yang, Dong and Roth, Holger and Xu, Daguang},
  journal={arXiv preprint arXiv:2201.01266},
  year={2022}
}

@inproceedings{ovadia2019can,
  title={Can you trust your model's uncertainty? Evaluating predictive uncertainty under dataset shift},
  author={Ovadia, Yaniv and others},
  booktitle={NeurIPS},
  year={2019}
}

@misc{CHAOSdata2019,
  author    = {Kavur, Ali Emre and Selver, M. Alper and Dicle, O{\u{g}}uz
               and Bar{\i}{\c{s}}, Mustafa and Gezer, N. Sinem},
  title     = {{CHAOS -- Combined (CT-MR) Healthy Abdominal Organ Segmentation Challenge Data}},
  year      = {2019},
  publisher = {Zenodo},
  doi       = {10.5281/zenodo.3362844},
  url       = {https://doi.org/10.5281/zenodo.3362844}
}

@ARTICLE{CAMUS,
  author={Leclerc, Sarah and Smistad, Erik and Pedrosa, João and Østvik, Andreas and Cervenansky, Frederic and Espinosa, Florian and Espeland, Torvald and Berg, Erik Andreas Rye and Jodoin, Pierre-Marc and Grenier, Thomas and Lartizien, Carole and D’hooge, Jan and Lovstakken, Lasse and Bernard, Olivier},
  journal={IEEE Transactions on Medical Imaging}, 
  title={Deep Learning for Segmentation Using an Open Large-Scale Dataset in 2D Echocardiography}, 
  year={2019},
  volume={38},
  number={9},
  pages={2198-2210},
  doi={10.1109/TMI.2019.2900516}}

@misc{yu2026agenticmemorylearningunified,
      title={Agentic Memory: Learning Unified Long-Term and Short-Term Memory Management for Large Language Model Agents}, 
      author={Yi Yu and Liuyi Yao and Yuexiang Xie and Qingquan Tan and Jiaqi Feng and Yaliang Li and Libing Wu},
      year={2026},
      eprint={2601.01885},
      archivePrefix={arXiv},
      primaryClass={cs.CL},
      url={https://arxiv.org/abs/2601.01885}, 
}

@misc{behrouz2024titanslearningmemorizetest,
      title={Titans: Learning to Memorize at Test Time}, 
      author={Ali Behrouz and Peilin Zhong and Vahab Mirrokni},
      year={2024},
      eprint={2501.00663},
      archivePrefix={arXiv},
      primaryClass={cs.LG},
      url={https://arxiv.org/abs/2501.00663}, 
}

@misc{xu2025amem,
      title={A-MEM: Agentic Memory for LLM Agents}, 
      author={Wujiang Xu and Zujie Liang and Kai Mei and Hang Gao and Juntao Tan and Yongfeng Zhang},
      year={2025},
      eprint={2502.12110},
      archivePrefix={arXiv},
      primaryClass={cs.CL},
      url={https://arxiv.org/abs/2502.12110}, 
}

@misc{zhang2025peft,
      title={Parameter-Efficient Fine-Tuning for Foundation Models}, 
      author={Dan Zhang and Tao Feng and Lilong Xue and Yuandong Wang and Yuxiao Dong and Jie Tang},
      year={2025},
      eprint={2501.13787},
      archivePrefix={arXiv},
      primaryClass={cs.CL},
      url={https://arxiv.org/abs/2501.13787}, 
}

@misc{zhuang2025foundationmodelmeetsfederated,
      title={When Foundation Model Meets Federated Learning: Motivations, Challenges, and Future Directions}, 
      author={Weiming Zhuang and Chen Chen and Jingtao Li and Chaochao Chen and Yaochu Jin and Lingjuan Lyu},
      year={2025},
      eprint={2306.15546},
      archivePrefix={arXiv},
      primaryClass={cs.LG},
      url={https://arxiv.org/abs/2306.15546}, 
}

@article{kavur2021chaos,
  title={CHAOS challenge-combined (CT-MR) healthy abdominal organ segmentation},
  author={Kavur, A Emre and Gezer, N Sinem and Bar{\i}{\c{s}}, Mustafa and Aslan, Sinem and Conze, Pierre-Henri and Groza, Vladimir and Pham, Duc Duy and Chatterjee, Soumick and Ernst, Philipp and {\"O}zkan, Sava{\c{s}} and others},
  journal={Medical image analysis},
  volume={69},
  pages={101950},
  year={2021},
  publisher={Elsevier}
}

@article{bernard2018deep,
  title={Deep learning techniques for automatic MRI cardiac multi-structures segmentation and diagnosis: is the problem solved?},
  author={Bernard, Olivier and Lalande, Alain and Zotti, Clement and Cervenansky, Frederick and Yang, Xin and Heng, Pheng-Ann and Cetin, Irem and Lekadir, Karim and Camara, Oscar and Ballester, Miguel Angel Gonzalez and others},
  journal={IEEE transactions on medical imaging},
  volume={37},
  number={11},
  pages={2514--2525},
  year={2018},
  publisher={ieee}
}

@article{leclerc2019deep,
  title={Deep learning for segmentation using an open large-scale dataset in 2D echocardiography},
  author={Leclerc, Sarah and Smistad, Erik and Pedrosa, Joao and {\O}stvik, Andreas and Cervenansky, Frederic and Espinosa, Florian and Espeland, Torvald and Berg, Erik Andreas Rye and Jodoin, Pierre-Marc and Grenier, Thomas and others},
  journal={IEEE transactions on medical imaging},
  volume={38},
  number={9},
  pages={2198--2210},
  year={2019},
  publisher={IEEE}
}

@misc{yang2023cardiacUDA,
      title={GraphEcho: Graph-Driven Unsupervised Domain Adaptation for Echocardiogram Video Segmentation}, 
      author={Jiewen Yang and Xinpeng Ding and Ziyang Zheng and Xiaowei Xu and Xiaomeng Li},
      year={2023},
      eprint={2309.11145},
      archivePrefix={arXiv},
      primaryClass={cs.CV},
      url={https://arxiv.org/abs/2309.11145}, 
}

@misc{wang2025parameterefficientfinetuninglargemodels,
      title={Parameter-Efficient Fine-Tuning in Large Models: A Survey of Methodologies}, 
      author={Luping Wang and Sheng Chen and Linnan Jiang and Shu Pan and Runze Cai and Sen Yang and Fei Yang},
      year={2025},
      eprint={2410.19878},
      archivePrefix={arXiv},
      primaryClass={cs.CL},
      url={https://arxiv.org/abs/2410.19878}, 
}

@article{simeoni2025dinov3,
  title={Dinov3},
  author={Sim{\'e}oni, Oriane and Vo, Huy V and Seitzer, Maximilian and Baldassarre, Federico and Oquab, Maxime and Jose, Cijo and Khalidov, Vasil and Szafraniec, Marc and Yi, Seungeun and Ramamonjisoa, Micha{\"e}l and others},
  journal={arXiv preprint arXiv:2508.10104},
  year={2025}
}

@article{gonzalez2023lifelongnnunet,
  title   = {Lifelong nnU-Net: A Framework for Standardized Medical Continual Learning},
  author  = {Gonz{\'a}lez, Camila and Ranem, Amin and Pinto dos Santos, Daniel
             and Othman, Ahmed and Mukhopadhyay, Anirban},
  journal = {Scientific Reports},
  volume  = {13},
  pages   = {9381},
  year    = {2023},
  doi     = {10.1038/s41598-023-34484-2}
}

@inproceedings{ranem2024unclesam,
  title     = {{UnCLe SAM}: Unleashing {SAM}'s Potential for Continual
               Prostate {MRI} Segmentation},
  author    = {Ranem, Amin and Aflal, Mohamed Afham Mohamed and
               Fuchs, Moritz and Mukhopadhyay, Anirban},
  booktitle = {Proceedings of Medical Imaging with Deep Learning},
  series    = {Proceedings of Machine Learning Research},
  volume    = {250},
  pages     = {1207--1220},
  year      = {2024}
}

@inproceedings{chen2024each,
  title={Each test image deserves a specific prompt: Continual test-time adaptation for 2d medical image segmentation},
  author={Chen, Ziyang and Pan, Yongsheng and Ye, Yiwen and Lu, Mengkang and Xia, Yong},
  booktitle={2024 IEEE/CVF Conference on Computer Vision and Pattern Recognition (CVPR)},
  pages={11184--11193},
  year={2024},
  organization={IEEE}
}

@inproceedings{ranem2025ncadapt,
  title={NCAdapt: Dynamic adaptation with domain-specific Neural Cellular Automata for continual hippocampus segmentation},
  author={Ranem, Amin and Kalkhof, John Orlando and Mukhopadhyay, Anirban},
  booktitle={Proceedings of the Winter Conference on Applications of Computer Vision},
  pages={3834--3843},
  year={2025}
}

@inproceedings{wang2026continual,
  title={Continual alignment for sam: Rethinking foundation models for medical image segmentation in continual learning},
  author={Wang, Jiayi and Dai, Wei and Wang, Haoyu and Yang, Sihan and Bi, Haixia and Sun, Jian},
  booktitle={Proceedings of the IEEE/CVF Conference on Computer Vision and Pattern Recognition},
  pages={7520--7529},
  year={2026}
}

@inproceedings{liu2024fedfms,
  title={Fedfms: Exploring federated foundation models for medical image segmentation},
  author={Liu, Yuxi and Luo, Guibo and Zhu, Yuesheng},
  booktitle={International conference on medical image computing and computer-assisted intervention},
  pages={283--293},
  year={2024},
  organization={Springer}
}

@article{zhang2025fedsca,
  title={Fedsca: federated tuning with similarity-guided collaborative aggregation for heterogeneous medical image segmentation},
  author={Zhang, Yumin and Gao, Yan and Duan, Haoran and Guo, Hanqing and Shah, Tejal and Ranjan, Rajiv and Wei, Bo},
  journal={arXiv preprint arXiv:2503.15390},
  year={2025}
}

@article{mcclelland1995complementary,
  title={Why There Are Complementary Learning Systems in the Hippocampus and Neocortex: Insights from the Successes and Failures of Connectionist Models of Learning and Memory},
  author={McClelland, James L. and McNaughton, Bruce L. and O'Reilly, Randall C.},
  journal={Psychological Review},
  volume={102},
  number={3},
  pages={419--457},
  year={1995},
  doi={10.1037/0033-295X.102.3.419}
}

@article{kumaran2016learning,
  title={What Learning Systems Do Intelligent Agents Need? Complementary Learning Systems Theory Updated},
  author={Kumaran, Dharshan and Hassabis, Demis and McClelland, James L.},
  journal={Trends in Cognitive Sciences},
  volume={20},
  number={7},
  pages={512--534},
  year={2016},
  doi={10.1016/j.tics.2016.05.004}
}

@article{wilson1994reactivation,
  title={Reactivation of Hippocampal Ensemble Memories During Sleep},
  author={Wilson, Matthew A. and McNaughton, Bruce L.},
  journal={Science},
  volume={265},
  number={5172},
  pages={676--679},
  year={1994},
  doi={10.1126/science.8036517}
}

@article{carr2011hippocampal,
  title={Hippocampal Replay in the Awake State: A Potential Substrate for Memory Consolidation and Retrieval},
  author={Carr, Margaret F. and Jadhav, Shantanu P. and Frank, Loren M.},
  journal={Nature Neuroscience},
  volume={14},
  number={2},
  pages={147--153},
  year={2011},
  doi={10.1038/nn.2732}
}

@article{nader2009single,
  title={A Single Standard for Memory: The Case for Reconsolidation},
  author={Nader, Karim and Hardt, Oliver},
  journal={Nature Reviews Neuroscience},
  volume={10},
  number={3},
  pages={224--234},
  year={2009},
  doi={10.1038/nrn2590}
}

@article{diekelmann2010memory,
  title={The Memory Function of Sleep},
  author={Diekelmann, Susanne and Born, Jan},
  journal={Nature Reviews Neuroscience},
  volume={11},
  number={2},
  pages={114--126},
  year={2010},
  doi={10.1038/nrn2762}
}

@article{ayzenberg2025protosam,
  title={ProtoSAM for automated one shot medical image segmentation using foundational models},
  author={Ayzenberg, Lev and Giryes, Raja and Greenspan, Hayit},
  journal={Scientific Reports},
  volume={15},
  number={1},
  pages={41482},
  year={2025},
  publisher={Nature Publishing Group UK London}
}

@article{li2025dino,
  title={Dino u-net: Exploiting high-fidelity dense features from foundation models for medical image segmentation},
  author={Li, Haoyue and Gao, Yifan and Yuan, Feng and Wang, Xiaosong and Gao, Xin},
  journal={arXiv preprint arXiv:2508.20909},
  year={2025}
}

\end{document}